%% file: main.tex
\documentclass[10pt,twocolumn,letterpaper]{article}

\usepackage[pagebackref=true,breaklinks=true,letterpaper=true,colorlinks,bookmarks=false]{hyperref}

\usepackage{iccv}
\usepackage{times}
\usepackage{epsfig}
\usepackage{graphicx}
\usepackage{amsmath}
\DeclareMathOperator*{\argmin}{argmin}
\usepackage{amssymb}

\usepackage{caption}
\usepackage{booktabs}
\usepackage{gensymb}
\usepackage[normalem]{ulem}
\usepackage{bbding}

\newcommand{\copyrightgoogle}{\href{https://www.google.com/help/terms_maps/}{\textcopyright 2023 Google}\xspace}

\usepackage[capitalize]{cleveref}
\crefname{section}{Sec.}{Secs.}
\Crefname{section}{Section}{Sections}
\Crefname{table}{Table}{Tables}
\crefname{table}{Tab.}{Tabs.}

\input{macros}

 \iccvfinalcopy 

\def\iccvPaperID{5711} 

\ificcvfinal\fi

\begin{document}

\title{AssetField: Assets Mining and Reconfiguration \\ in \Representation}
\author{Yuanbo Xiangli$^1$$^*$ , Linning Xu$^1$$^*$ , Xingang Pan$^{3}$, Nanxuan Zhao$^{4}$, Bo Dai$^{2}$\Envelope, Dahua Lin$^{1,2}$\\
	$^{1}$ The Chinese University of Hong Kong \quad
	$^{2}$ Shanghai AI Laboratory \\
	$^{3}$ Max Planck Institute for Informatics \quad 
	$^{4}$ Adobe Research \\
	{\tt\small \{xy019,xl020,dhlin\}@ie.cuhk.edu.hk}~
	{\tt\small xpan@mpi-inf.mpg.de}~ \\
	{\tt\small nanxuanzhao@gmail.com}~
	{\tt\small daibo@pjlab.org.cn}
}


\twocolumn[{%
	\renewcommand
	\twocolumn[1][]{#1}%
	\maketitle
	\begin{center}
		\centering
		\vspace{-25pt}
		\includegraphics[width=0.95\textwidth]{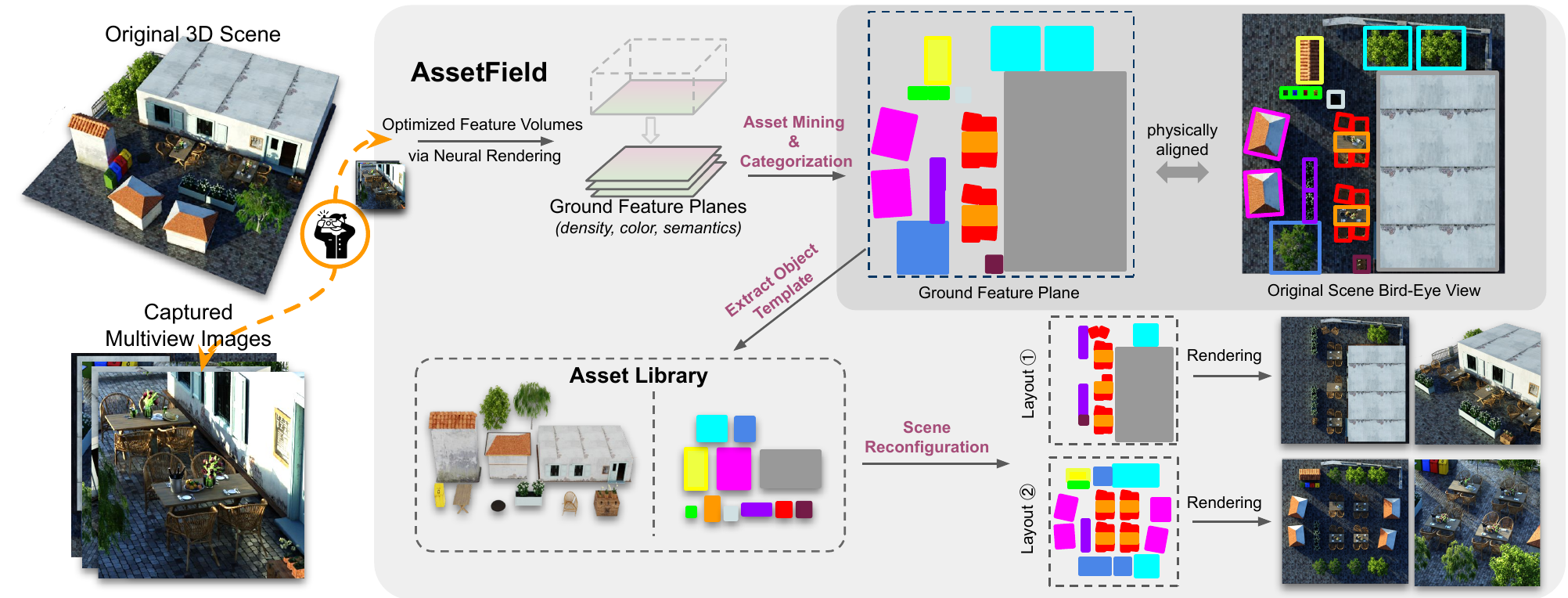}
		\vspace{-10pt}
		\captionof{figure}{\small
			Man-made environments are often characterized by repetitive scene objects,~\eg tables, chairs, and trees. \emph{AssetField} represents these environments with a set of informative ground feature planes aligning with the physical ground, from which neural representations of scene objects are extracted and grouped into categories. The proposed mechanism allows users to manipulate and compose assets directly on the ground feature plane and produces high-quality rendering on novel scene configurations.
		}
		\vspace{-5pt}
		\label{fig:teaser}
	\end{center}
}]

\ificcvfinal\thispagestyle{empty}\fi

\input{sections/0_abs.tex}
\input{sections/1_intro.tex}
\input{sections/2_related.tex}
\input{sections/3_method.tex}

\input{sections/4_exp.tex}
\input{sections/5_conclusion.tex}

{\small
\bibliographystyle{ieee_fullname}
\bibliography{egbib}
}

\end{document}

%% file: macros.tex
\usepackage{xcolor}

\newcommand{\Representation}{Ground Feature Plane Representation}

\newcommand{\tb}[1]{{\textbf{#1}}}

\usepackage{soul}

%% file: sections/0_abs.tex
\begin{abstract}
\label{sec:abs}

\vspace{-20pt}
Both indoor and outdoor environments are inherently structured and repetitive. 
Traditional modeling pipelines keep an asset library storing unique object templates, which is both versatile and memory efficient in practice. 
Inspired by this observation, 
we propose AssetField, 
a novel neural scene representation that learns a set of object-aware ground feature planes to represent the scene, where an asset library storing template feature patches can be constructed in an unsupervised manner.
Unlike existing methods which require object masks to query spatial points for object editing, our ground feature plane representation offers a natural visualization of the scene in the bird-eye view, allowing a variety of operations~(\eg translation, duplication, deformation) on objects to configure a new scene. With the template feature patches, group editing is enabled for scenes with many recurring items to avoid repetitive work on object individuals. 
We show that AssetField not only achieves competitive performance for novel-view synthesis but also generates realistic renderings for new scene configurations.

\vspace{-10pt}
\end{abstract}

%% file: sections/1_intro.tex

\section{Introduction}
\label{sec:intro}
\vspace{-5pt}
The demand for bringing our living environment into a virtual realm continuous to increase these days, with example cases ranging from indoor scenes such as rooms and restaurants, to outdoor ones like streets and neighborhoods.
Apart from the realistic 3D rendering, real-world applications also require flexible and user-friendly editing of the scene.
Use cases can be commonly found in interior design, urban planning etc. To save human labor and expense, users need to frequently visualize different scene configurations before finalizing a plan and bringing it to reality, like shown in Fig.\ref{fig:teaser}.
For their interests, a virtual environment offering versatile editing choices and high rendering quality is always preferable.
In these scenarios, objects are primarily located on a horizontal plane like ground, and can be inserted to/deleted from the scene.
Translation along the plane and rotation around the vertical axis are also common operations. 
Furthermore, group editing becomes essential when scenes are populated with recurring items~(\eg substitute all chairs with stools and remove all sofas in a bar).

While recent advances in neural rendering \cite{mildenhall2020nerf,Barron2021MipNeRF3U,Yu2022PlenoxelsRF,mueller2022instant}
offer promising solutions to producing realistic visuals, they struggle to meet the aforementioned editing demands.
Specifically, traditional neural radiance field (NeRF)-based methods such as~\cite{Zhang2020NeRFAA,MartinBrualla2020NeRFIT,Barron2021MipNeRF3U} encode an entire scene into a single neural network, making it difficult to manipulate and composite due to its implicit nature and limited network capacity.
Some follow-up works~\cite{Yang2022NeuralRI,Guo2020ObjectCentricNS} tackle object-aware scene rendering in a bottom-up fashion by learning one model per object and then performing joint rendering. 
Another branch of  methods learn object radiance fields using instance masks~\cite{yang2021objectnerf}, object motions~\cite{Yuan2021STaRST}, and image features~\cite{tschernezki22neural,kobayashi2022distilledfeaturefields} as clues but are scene-specific, limiting their applicable scenarios.
Recently, some approaches have attempted to combine voxel grids with neural radiance fields~\cite{Liu2020NeuralSV,Yu2022PlenoxelsRF,mueller2022instant} to explicitly model the scene. 
Previous work~\cite{Liu2020NeuralSV} showed local shape editing and scene composition abilities of the hybrid representation. 
However, since the learned scene representation is not object-aware, users must specify which voxels are affected to achieve certain editing requirements, which is cumbersome, especially for group editing.
Traditional graphical workflows build upon an asset library that stores template objects, whose copies are deployed onto a `canvas' by designers, then rendered by some professional software~(\eg interior designers arrange furniture according to floorplans). 
This practice significantly saves memory for large scene development and offers users versatile editing choices,
which inspires us to resemble this characteristic in neural rendering. 

To this end, we present \emph{AssetField}, a novel neural representation that bears the editing flexibility of traditional graphical workflows.
Our method factorizes a 3D neural field into a ground feature plane and a vertical feature axis. As illustrated in Fig.~\ref{fig:teaser}, the learned \textbf{ground feature plane} is a 2D feature plane that is visually aligned with the bird-eye view (BEV) of the scene, allowing intuitive manipulation of individual objects. 
It is also able to embed multiple scenes into scene-specific ground feature planes with a shared vertical feature axis, rendered using a shared MLP.
The learned ground feature planes encode scene density, color and semantics, providing rich clues for object detection and categorization.
We show that assets mining and categorization, and scene layout estimation can be directly performed on the ground feature planes. 
By maintaining a cross-scene \emph{asset library} that stores template objects’ ground feature patches, our method enables versatile editing at \emph{object-level}, \emph{category-level}, and \emph{scene-level}.

In summary, AssetField
1) learns a set of explicit ground feature planes that are intuitive and user-friendly for scene manipulation;
2) offers a novel way to discover assets and scene layout on the informative ground feature planes, from which one can construct an asset library storing feature patches of object templates from multiple scenes;
3) improves group editing efficiency and enables versatile scene composition and reconfiguration and
4) provides realistic renderings on new scene configurations.

%% file: sections/2_related.tex

\section{Related Works}
\label{sec:related}
\noindent\textbf{Neural Implicit Representations and Semantic Fields.}
Since the introduction of neural radiance fields~\cite{mildenhall2020nerf}, many advanced scene representations have been proposed\cite{liu2020neural,Yu2022PlenoxelsRF,mueller2022instant,Chen2022ECCV,liu2020neural,Chan2021,mueller2022instant}, 
demonstrating superior performance in terms of quality and speed for neural renderings. 
However, most of these methods are semantic and content agnostic, and many assume sparsity to design a more compact structure for rendering acceleration~\cite{liu2020neural,Chen2022ECCV,mueller2022instant}. We notice that the compositional nature of a scene and the occurrence of repetitive objects within can be further utilized, where we can extract a reusable asset library for more scalable usages, similar to those adopted in the classical modeling pipeline.

A line of recent neural rendering works has explored the jointly learning a semantic fields along with the original radiance field. Earlier works use available semantic labels \cite{zhi2021place} or existing 2D detectors for supervision~\cite{kundu2022panoptic}. The realized semantic field can enable category or object-level control. More recently, \cite{tschernezki22neural,kobayashi2022distilledfeaturefields} explore the potential of distilling self-supervised 2D image feature extractors~\cite{caron2021emerging,amir2021deep,fan2022nerf} into NeRF, and showcasing their usages of support local editing. In this work, we target an orthogonal editing goal where the accurate control of high-level scene configuration and easy editing on object instances is desired.

\smallskip
\noindent\textbf{Object Manipulation and Scene Composition.}
Traditional modeling and rendering pipelines~\cite{Bleyer2011PatchMatchS, Broadhurst2001APF, Schnberger2016StructurefromMotionR, Schnberger2016PixelwiseVS, Seitz1997PhotorealisticSR,Karsch2011RenderingSO} are vastly adopted for scene editing and novel view synthesis in early approaches. For example, Karsch~\etal~\cite{Karsch2011RenderingSO} propose to realistically insert synthetic objects into legacy images by creating a physical model of the scene from user annotations of geometry and lighting conditions, then compose and render the edited scene. Cossairt~\etal~\cite{Cossairt2008LightFT} consider synthetic and real objects compositions from the perspective of light field, where objects are captured by a specific hardware system.
\cite{zheng2012interactive,kim2012acquiring,kholgade20143d} consider the problem of manipulating existing 3D scenes by matching the objects to cuboid proxies/pre-captured 3D models.

These days, several works propose to tackle object-decomposite rendering under the context of newly emerged neural implicit representations~\cite{mildenhall2020nerf}. 
Ost~\etal~\cite{Ost2021NeuralSG} target dynamic scenes and learn a scene graph representation that encodes object transformation and radiance at each node, which further allows rendering novel views and re-arranged scenes. 
Kundu~\etal \cite{kundu2022panoptic} resort to existing 3D object detectors for foreground object extraction.
Sharma~\etal~\cite{sharma2022seeing} disentangles static and movable scene contents, leveraging object motion as a cue.
Guo~\etal~\cite{Guo2020ObjectCentricNS} propose to learn object-centric neural scattering functions to implicitly model per-object light transportation, enabling scene rendering with moving objects and lights. 
Neural Rendering in a Room~\cite{Yang2022NeuralRI} targets indoor scenes by learning a radiance field for each pre-captured object and putting objects into a panoramic image for optimization. 
While these methods need to infer object from motion, or require one model per object, ObjectNeRF~\cite{Yang2021LearningON} learns a decompositional neural radiance field, utilizing semantic masks to separate objects from the background to allow editable scene rendering.
uORF~\cite{yu2022unsupervised} performs unsupervised discovery of object radiance fields without the need for semantic masks, but requires cross-scene training and is only tested on simple synthetic objects without textures.

%% file: sections/3_method.tex
\section{AssetField}
\label{sec:method}

In this work, we primarily consider a branch of real-world application scenarios that require fast and high-quality rendering of scenes whose configuration is subject to change, such as interior design, urban planning and traffic simulation. 
In these cases, objects are mainly placed on some dominant horizontal plane, and is commonly manipulated with insertion, deletion, translation on the horizontal plane, and rotation around the vetical axis, etc.

We first introduce our ground feature plane representation in Sec.~\ref{subsec:ground_feature_plane} to model each neural field.
Sec.~\ref{subsec:asset_mining} describes the process of assets mining with the inferred the ground feature plane. 
We further leverage the color and semantic feature planes to categorize objects in an unsupervised manner, which is illustrated in Sec.~\ref{subsec:asset_grouping}. 
Finally, Sec.~\ref{subsec:asset_library} demonstrates the construction of a cross-scene asset library that enables versatile scene editing.

\subsection{Ground Feature Plane Representation}
\label{subsec:ground_feature_plane}
Ground plan has been commonly used for indoor and outdoor scene modeling~\cite{sharma2022seeing,devries2021unconstrained,Saha2022TranslatingII}.
We adopt a similar representation to parameterize a 3D neural field with a 2D ground feature plane $\mathcal{M}$ of shape $L\times W \times N$, and a globally encoded vertical feature axis $\mathcal{H}$ of shape $H \times N$, where $N$ is the feature dimension. A query point at coordinate $(x,y,z)$ is projected onto $\mathcal{M}$(plane) and $\mathcal{H}$(axis) to retrieve its feature values $m$ and $h$ via bilinear/linear interpolation:
\begin{equation}
		m=\operatorname{Interp}(\mathcal{M},(x,y)), 
		h=\operatorname{Interp}(\mathcal{H},z), 
\end{equation}	
which are then combined and decoded into the 3D scene feature via a MLP decoder.
Concretely, a 3D radiance field is parameterized by a set of ground feature planes $\mathcal{M}$=$(\mathcal{M}_\sigma, \mathcal{M}_{c})$, and vertical feature axes $\mathcal{H}$=$(\mathcal{H}_\sigma, \mathcal{H}_{c})$, for the density and color fields respectively.
The retrieved feature values $m$=$(m_{\sigma},m_{c})$ and $h$=$(h_{\sigma}, h_{c})$ are then combined and decoded into point density $\sigma$ and view-dependent color $c$ values by two decoders $Dec_\sigma$, $Dec_{rgb}$.
Points along a ray $\mathbf{r}$ are volumetrically integrated following~\cite{mildenhall2020nerf}:
\vspace{-5pt}
\begin{equation}
	\vspace{-5pt}
	\hat{C}(\mathbf{r})=\sum_{i=1}^N T_i\left(1-\exp \left(-\sigma_i \delta_i\right)\right) c_i,
	\label{eq:volumn-rendering}
\end{equation}
where $T_i=\exp (-\sum_{j=1}^{i-1} \sigma_j \delta_j)$, and supervised by the 2D image reconstruction loss with $\sum_{\mathbf{r}}(\|\hat{C}(\mathbf{r})-C(\mathbf{r})\|_2^2)$, where $C(\mathbf{r})$ is the ground truth pixel color. 

\begin{figure}
	\centering
	\includegraphics[width=0.9\linewidth]{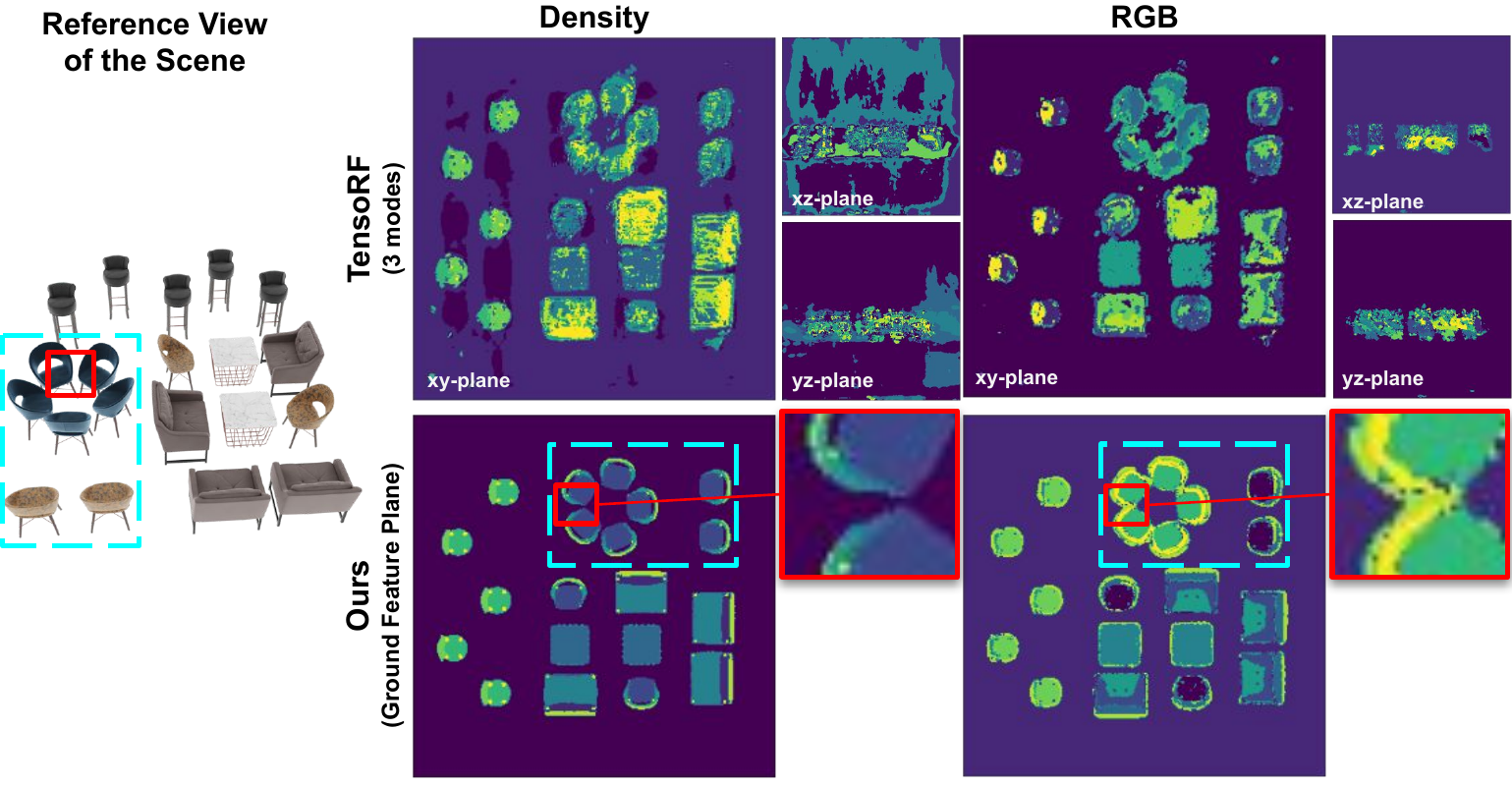}
	\vspace{-5pt}
	\caption{\small TensoRF with full $3D$ factorization produces noisy feature planes; our ground plane representation yields informative features that clearly illustrated scene contents and layout after discretization, especially in the density field. 
	{\color{red}Red} boxes: two spatially close objects can be clearly separated on the density plane but not the RGB plane. 
	{\color{cyan}Blue} boxes: objects with similar geometry but different appearance can be distinguished on the RGB plane but not the density plane.}
	\vspace{-15pt}
	\label{fig:tensorf_vs_groundplan}
\end{figure}

Such neural representation are beneficial to our scenario. 
Firstly, the ground feature planes are naturally aligned with the BEV of the scene, mirroring the human approach to high-level editing and graphic design, where artists and designers mainly sketch on 2D canvas to reflect a 3D scene. 
Secondly, the globally encoded vertical feature axis encourages the ground feature plane to encode more scene information, which aligns better with scene contents. 
Thirdly, this compact representation is more robust when trained with sparse view images, where the full 3D feature grids are easy to overfit under insufficient supervision, producing noisy values, as depicted in Fig.~\ref{fig:tensorf_vs_groundplan}. 

\begin{figure*}
	\centering
	\includegraphics[width=0.98\linewidth]{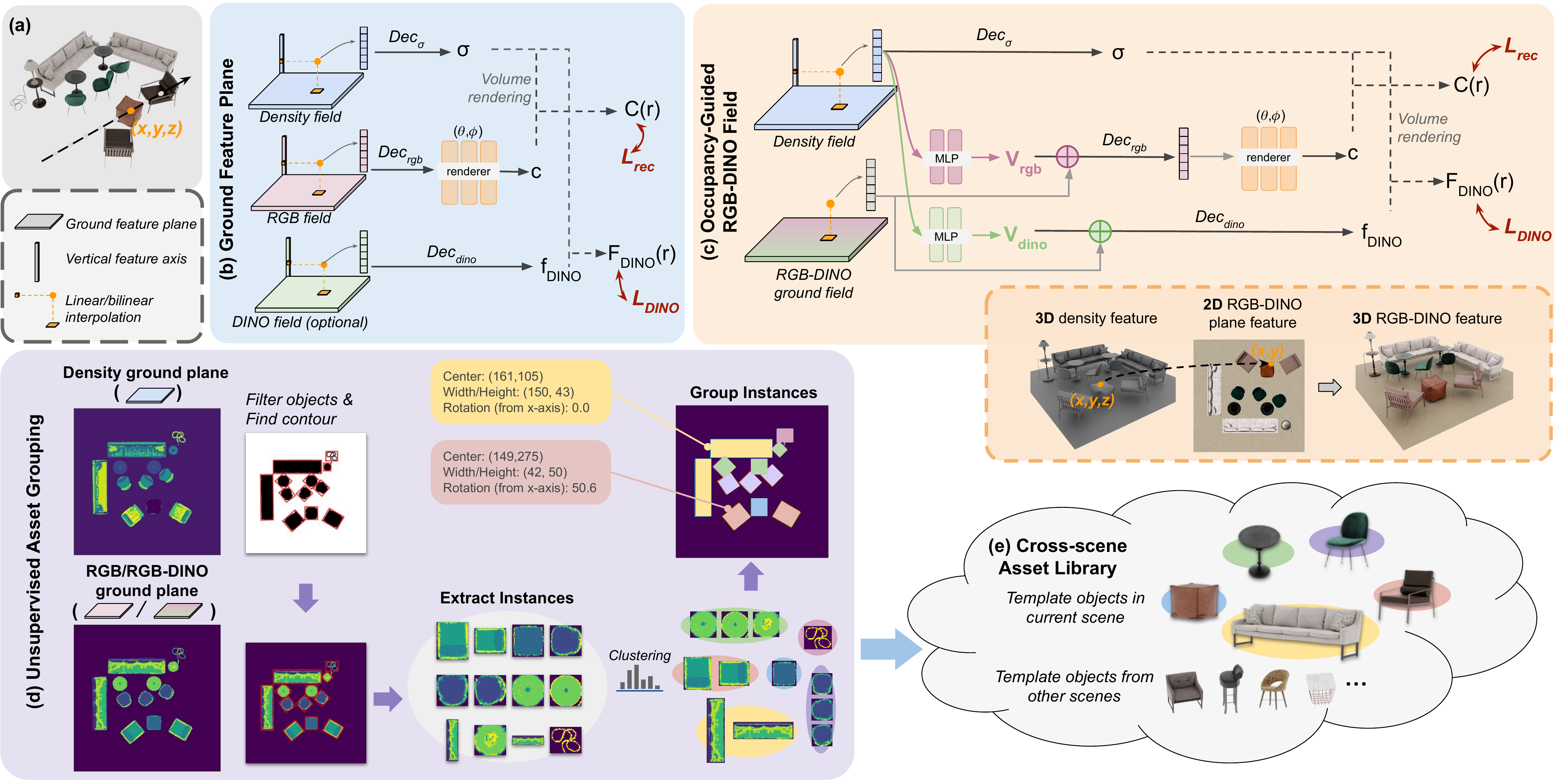}
	\vspace{-5pt}
	\caption{\small Overview of \emph{AssetField}. (a) We demonstrate on a scene without background for clearer visuals. (b) The proposed ground feature plane representation factorizes a neural field into a horizontal feature plane and a vertical feature axis. (c) We further integrate color and semantic field into a 2D neural plane, which is decoded into 3D-aware features with the geometry guidance from scene density. The inferred RGB-DINO plane is rich in object appearance and semantic clues whilst being less sensitive to vertical displacement between objects, on which we can (d) detect assets and grouping them into categories. (e) For each category, we select a template object and store its density and color ground feature patches into the asset library. A cross-scene asset library can be construct by letting different scenes fit there own ground feature planes whilst sharing the same vertical feature axes and decoders/renderers.}
	\vspace{-18pt}
	\label{fig:pipeline}
\end{figure*}

\subsection{Assets Mining on Ground Feature Plane}
\label{subsec:asset_mining}

\begin{figure}
	\centering
	\includegraphics[width=\linewidth]{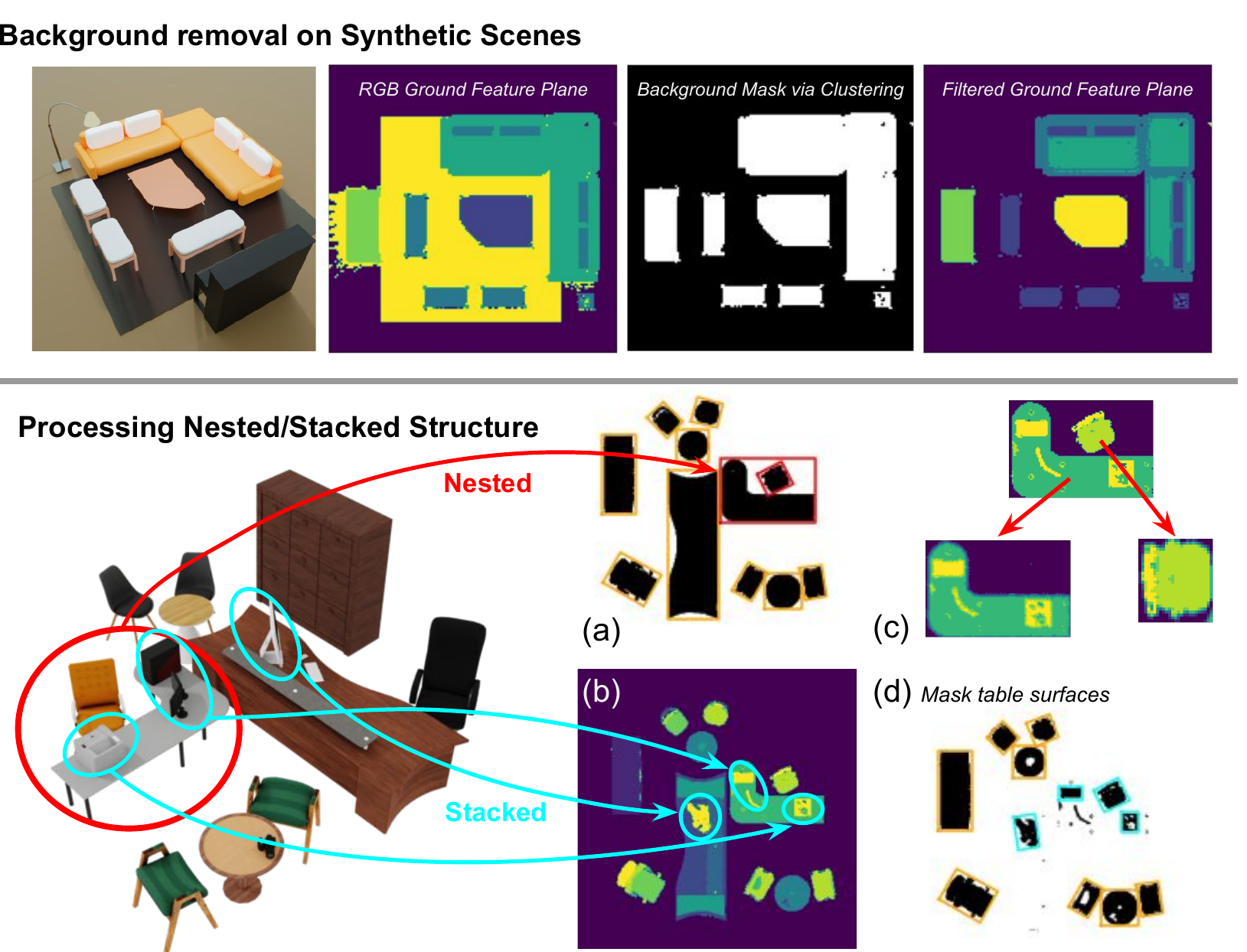}
	\vspace{-15pt}
	\caption{\small Top: Simple scene background can be filtered on the learned ground feature plane in advance using feature clustering. Bottom: (a) Nested structure can be separated by (c) firstly identify the enclosed chair, then set its value to background feature for table patch. (b) Items placed on top of a surface can be detected by (d) another round of filtering that treats table surface as background.}
	\vspace{-20pt}
	\label{fig:hierarchical_structure}
\end{figure}

For the ease of demonstration, let us first consider a simplified case where objects are scattered on an invisible horizontal plane, as in Fig.~\ref{fig:pipeline}~(a).
On scenes with a background, a pre-filtering step can be performed on the learned ground feature plane as illustrated in Fig.~\ref{fig:hierarchical_structure}.
We start from modeling the radiance field,
where a set of ground features planes $\mathcal{M}$=$(\mathcal{M}_\sigma, \mathcal{M}_{c})$ describing scene density and color are inferred following the formulation in Sec.~\ref{subsec:ground_feature_plane}.
It can be observed that $\mathcal{M}_\sigma$ tends to exhibit sharper object boundaries compared to the color feature plane, as shown in the red boxes in Fig.~\ref{fig:tensorf_vs_groundplan}.
This could be attributed to the mechanism of neural rendering (Eq.~\ref{eq:volumn-rendering}), where the model firstly learns a clean and accurate density field to guide the learning of the color field. 
We therefore prefer to use $\mathcal{M}_\sigma$ for assets mining.
In the example scene, the feature plane is segmented into two clusters with K-means~\cite{Lloyd1982LeastSQ} to obtain a binary mask of the objects.
Contour detection~\cite{Suzuki1985TopologicalSA,opencv_library} is then applied to locate each object, resulting in a set of bounding box. 
Note that the number of clusters can be customized according to the objects users want to highlight. 
In more complex scenarios where objects are arranged in a hierarchical structure,~(\eg computer - table - floor), the clustering step can be repeated to gradually unpack the scene, as illustrated in Fig.~\ref{fig:hierarchical_structure}. 
With the bounding boxes, a collection of object neural representations $\mathcal{P}$=$\{(p_{\sigma}, p_{c})\}$ can be obtained, which are the enclosed feature patches on $\mathcal{M}_\sigma$ and $\mathcal{M}_{rgb}$. 
To address complex real-world scenes, we take inspiration from previous works ~\cite{kobayashi2022distilledfeaturefields,tschernezki22neural} that models a DINO~\cite{Caron2021EmergingPI} field to guide the learning a semantic-aware radiance field.
Similarly, we can learn a \emph{separate} DINO ground feature plane $\mathcal{M}_{dino}$ to provide more explicit indications of object presence. 
As AssetField models a set of separate fields, object discovery can be conducted on any field that offers the most distinctive features in a scene dependent manner.

At this point, users can intuitively edit the scene with object feature patches $\mathcal{P}$, \eg,``paste" $(p_{\sigma}^{i}$, $p_{c}^{i})\in\mathcal{P}$ to the designated location on $\mathcal{M}_\sigma$ and $\mathcal{M}_{rgb}$ to insert object $i$. The edited ground feature planes $\mathcal{M'}$=$(\mathcal{M'}_\sigma, \mathcal{M'}_{c})$ get paired with the original vertical feature axes $\mathcal{H}$=$(\mathcal{H}_\sigma, \mathcal{H}_{c})$ to decode 3D information using the original $Dec_\sigma$ and $Dec_{rgb}$.

\subsection{Unsupervised Asset Grouping}
\label{subsec:asset_grouping}
Despite being versatile, users can only interact with individual instances in $\mathcal{P}$ from the learned ground planes, whereas group editing is also a desirable feature in real-world applications, especially when objects of the same category need to be altered together. While the definition of object category can be narrow or broad, here we assume that objects with close appearance and semantics are analogues and use RGB and semantic feature plane for assets grouping. A case where the density features fail to distinguish two visually different objects is highlight in Fig.~\ref{fig:tensorf_vs_groundplan}.

\noindent \textbf{Occupancy-Guided RGB-DINO field.} 
As our goal is to ``self-discover" assets from neural scene representation, there is no extra prior on object category to regularize scene features. 
3D voxel-based methods such as those described in ~\cite{liu2020neural,Yu2022PlenoxelsRF}, may learn different sets of features to express the same objects, as grid features are independently optimized. 
Such issue can be alleviated by our proposed neural representation, where the ground feature plane $\mathcal{M}$ is constrained by the globally shared vertical feature axis $\mathcal{H}$. 
Concretely, given two identical objects $i,j$ placed on a horizontal flat surface, the same feature chunk on $\mathcal{H}$ will be queried during training, which constraints their corresponding feature patches $p_i$ and $p_j$ to be as similar as possible so that they can be decoded into the same set of 3D features.
However, such constraint no longer holds when there is a vertical displacement among identical objects~(\eg one the ground and one on the table), where different feature chunks on $\mathcal{H}$ are queried, leading to divergent $p_i$ and $p_j$.

To learn a more object-centric ground feature plane rich in color and semantics clues, 
we propose to integrate the color and semantic fields by letting them share the same set of ground feature planes, denoted by $\mathcal{M}_{rgb{\text -}dino}$. 
Instead of appending a vertical feature axis, here we use scene density features to guide the decoding of $\mathcal{M}_{rgb{\text -}dino}$ into 3D-aware features, as illustrated in Fig.~\ref{fig:pipeline}(c).
It can be interpreted as $\mathcal{M}_\sigma$ and $\mathcal{H}_\sigma$ fully capture the scene geometry, while $\mathcal{M}_{rgb{\text -}dino}$ captures the `floorplan' of scene semantics layouts and appearances. 
For a query point at $(x,y,z)$, its retrieved density feature $m_\sigma$ and $h_\sigma$ are mapped to a color feature $v_{rgb}$ and a semantic feature $v_{dino}$ via two MLPs, 
which are then decoded into scene color $c$ and semantic $f_{dino}$ along with the RGB-DINO plane feature $m_{rgb{\text -}dino}=\operatorname{Interp}(\mathcal{M}_{rgb{\text -}dino},(x,y))$ via $Dec_{rgb}$ and $Dec_{dino}$.

\smallskip
\noindent \textbf{Assets Grouping and Template Matching.} 
On the inferred RGB-DINO ground feature plane, we then categorize the discovered objects by comparing their RGB-DINO feature patches enclosed in bounding boxes. 
However, due to the absence of object pose information, pixel-wise comparison is not ideal. 
Instead, we compare the distributions of color and semantic features among patches. 
To do this, we first discretize $\mathcal{M}_{rgb{\text -}dino}$ with clustering~(\eg K-means), 
which results in a set of labeled object feature patches $\mathcal{K}$. 
The similarity between two object patches $k_i,k_j\in\mathcal{K}$ are measured by the Jensen-Shannon Divergence over the distribution of labels, denoted by $\operatorname{JSD}(k_i||k_j)$.
Agglomerative clustering~\cite{Mllner2011ModernHA} is then performed using JS-divergence as the distance metric. The number of clusters can be set by inspecting training views, and can be flexibly adjusted to fit users' desired categorization granularity. 

With scene assets grouped into categories, a \emph{template} object can be selected from each cluster either randomly or in a user-defined manner.
We can further extract scene layout in BEV by computing the relative pose between the template object and its copies,~\ie to optimize a rotation angle $\theta$ that minimizes the pixel-wise loss between the RGB-DINO feature patches of the template and each copy with
$\theta^{\ast} = \argmin_{\theta} \sum_{i}^{N}||\hat{p}_i - R_\theta(p)_i||^2_2$ for $p\in\mathcal{P}_{rgb{\text -}dino}$,
where $\hat{p}$ is the template RGB-DINO feature patch, $R_\theta$ rotates the input feature patch by $\theta$.

\subsection{Cross-scene Asset Library}
\label{subsec:asset_library}
Following the proposed framework, a scene can be represented with (1) a set of template feature patches $\mathcal{P}$=$\{(p_\sigma,p_{rgb})\}$, (2) a layout describing object position and pose in the BEV, (3) the shared vertical feature axes $\mathcal{H}=(\mathcal{H}_\sigma,\mathcal{H}_{rgb})$, and (4) MLP decoders $Dec_\sigma$, $Dec_{rgb}$, 
which enables versatile scene editing at object-, category-, and scene-level. The newly configured scenes can be directly rendered without retraining. 	
An optional template refinement step is also allowed. Examples are given in Sec.~\ref{sec:exp}.

Previous work~\cite{liu2020neural} demonstrates that voxel-based neural representations support multi-scene modeling by learning different voxel embeddings for each scene whilst sharing the same MLP renderer.
However, it does not support cross-scene analogue discovery due to the aforementioned lack of constraints issue, 
whereas in reality, objects are not exclusive to a scene.
Our proposed neural representation has such potential to discover cross-scene analogues by also sharing the vertical feature axes among different scenes. 
Consequently, we can construct a cross-scene asset library storing template feature patches, and continuously expand it to accommodate new ones.

%% file: sections/4_exp.tex

\section{Experiment}
\label{sec:exp}

In this section, we first describe our experiment setup, then evaluate AssetField on novel view synthesis both quantitatively and qualitatively, demonstrating its advantages in asset mining, categorization, and editing flexibility.
More training details and ablating results of hyper-parameters~(\eg the number of clusters, the pairing of plane feature, and axis feature) are provided in supplementary.

\subsection{Experimental Setup}
\noindent \textbf{Dataset.}
A synthetic dataset is created for evaluation. We compose $10$ scenes resembling common man-made environments such as conference room, living room, dining hall and office. Each scene contains objects from 3$\sim$12 categories with a fixed light source. For each scene, we render $50$ views with viewpoints sampled on a half-sphere, among which $40$ are used for training and the rest for testing. 
We demonstrate flexible scene manipulation with AssetField on both the synthetic and real-world data, including scenes from Mip-NeRF 360~\cite{barron2022mip}, DONeRF~\cite{neff2021donerf}, and ObjectNeRF~\cite{yang2021objectnerf}. We also show manipulation results on city scenes collected from Google Earth Studio~\cite{google_earth_studio}.

\noindent \textbf{Implementation.}
We use NeRF~\cite{mildenhall2020nerf} and TensoRF~\cite{Chen2022ECCV} as baselines to evaluate the rendering quality of the original scenes.
For a fair comparison, all methods are implemented to model an additional DINO field. Specifically,
(1) NeRF is extended with an extra head to predict view-independent DINO feature~\cite{amir2021deep} in parallel with density.
(2) For TensoRF, we additionally construct the DINO field which is factorized along $3$ directions the same as its radiance field.
(3) \textbf{S(tandard)-AssetField} separately models the density, RGB, and DINO fields.
(4) \textbf{I(ntegrated)-AssetField }models the density field the same as S-AssetField, and an integrated RGB-DINO ground feature plane. 
Both S-AssetField and I-AssetField adopt outer-product to combine ground plane features and vertical axis features, following~\cite{Chen2022ECCV}.
The resolution of feature planes in TensoRF baseline and AssetField are set to 300$\times$300.
Detailed model adaptation can be found in the supplementary.
We train NeRF for $200k$ iterations, and $50k$ iterations for TensoRF and AssetField using Adam~\cite{Kingma2015AdamAM} optimization with a learning rate set to $5e^{-4}$ for NeRF and $0.02$ for TensoRF and AssetField.

\subsection{Results}
\begin{table}[t!]
	\begin{center}
		\resizebox{\linewidth}{!}{
			\begin{tabular}{l|rrr|rrr|rrr|rrr}
				\toprule
				&   \multicolumn{3}{c|}{\emph{Scene1}}  &   \multicolumn{3}{c|}{\emph{Scene2}} &   \multicolumn{3}{c|}{\emph{Scene3}} & \multicolumn{3}{c}{\emph{Scene4}} \\
				&  PSNR & SSIM & LPIPS  &
				PSNR & SSIM & LPIPS &
				PSNR & SSIM & LPIPS  & PSNR & SSIM  & LPIPS   \\
				\midrule
				NeRF   & 32.977 & 0.969 & 0.067  & 35.743 & 0.967 & 0.051 & 32.521 & 0.959 & 0.058 & 34.212 & 0.964 & 0.072 \\
				TensoRF   & 35.751 & 0.990 & 0.057 & \tb{38.184} & \tb{0.995} & \tb{0.027} & \underline{36.933} & \underline{0.994} & \underline{0.034} & \tb{37.795} & \tb{0.993} & \tb{0.059}\\
				\midrule
				S-AssetField & 36.471 & 0.992 & 0.049 & 36.856 &  0.993  & 0.037 & 36.753 & 0.994 & 0.038 & 37.445 & 0.990 & 0.065 \\
				I-AssetField & \tb{36.526} & \tb{0.992} & \tb{0.047} & \underline{37.271} & \underline{0.994} & \underline{0.035} & \tb{37.249} & \tb{0.995} & \tb{0.032} & \underline{37.716} & \underline{0.991} & \underline{0.060} \\
				\bottomrule
			\end{tabular}		
		}
		\vspace{-15pt}
	\end{center}
	\centering
	\caption{\small Quantitative comparison on test views for the $4$ scenes in Fig.~\ref{fig:editing_results}. We report PSNR($\uparrow$), SSIM($\uparrow$)~\cite{Wang2004ImageQA} and LPIPS($\downarrow$)~\cite{Zhang2018TheUE} for evalution. The \tb{best} and \underline{second best} results are highlighted.}
	\vspace{-15pt}
	\label{tab:recon}
\end{table}

\noindent \textbf{Novel View Rendering.}
We compare S-AssetField and I-AssetField with the adapted NeRF~\cite{mildenhall2020nerf} and TensoRF~\cite{Chen2022ECCV} as described above. 
Quantitative results are provided in Tab.~\ref{tab:recon}. 
It is noticeable that AssetField's ground feature plane representation~(\ie $xy{\text -}z$) achieves comparable performance with TensoRF's $3$-mode factorization~(\ie $xy{\text -}z$, $xz{\text -}y$,$yz{\text -}x$), indicating the suitability of adopting ground plane representations for such scenes. Our method also inherits the merit of efficient training and rendering from grid-based methods. Compared to NeRF, our model converges 40x faster at training and renders 30x faster at inference.

\begin{figure}
	\centering
	\includegraphics[width=\linewidth]{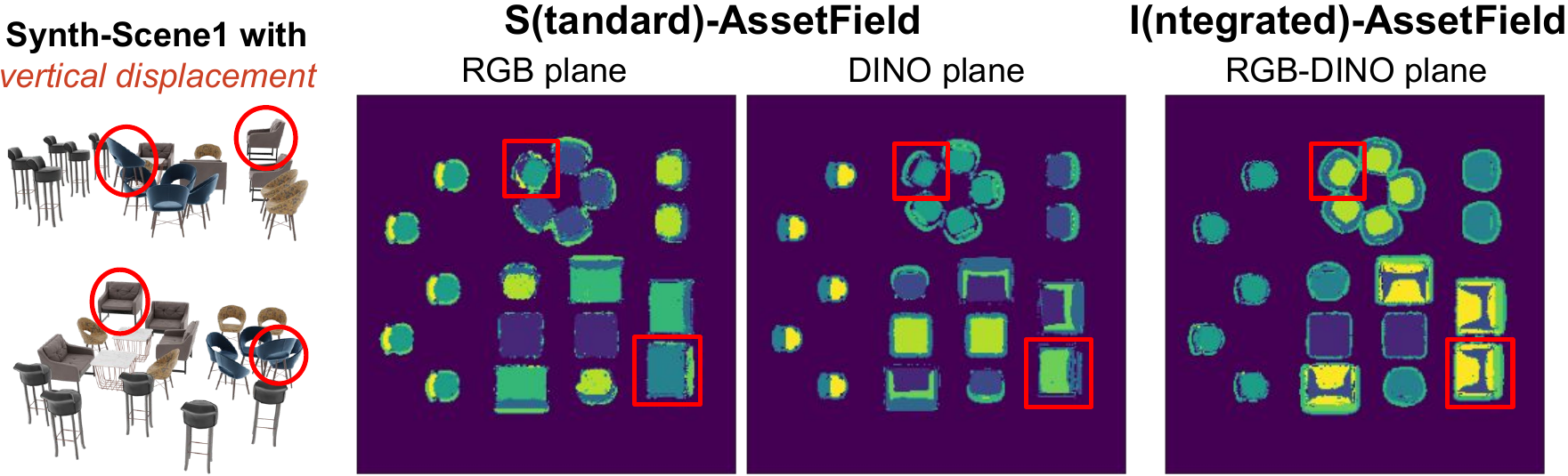}
	\caption{\small The RGB-DINO ground feature plane from I-AssetField yields consistent features for analogues with vertical displacement, whereas S-AssetField infers different set of features due to the lack of constraints.}
	\label{fig:occupancy-guided}
	\vspace{-10pt}
\end{figure}

\begin{figure}
	\centering
	\includegraphics[width=\linewidth]{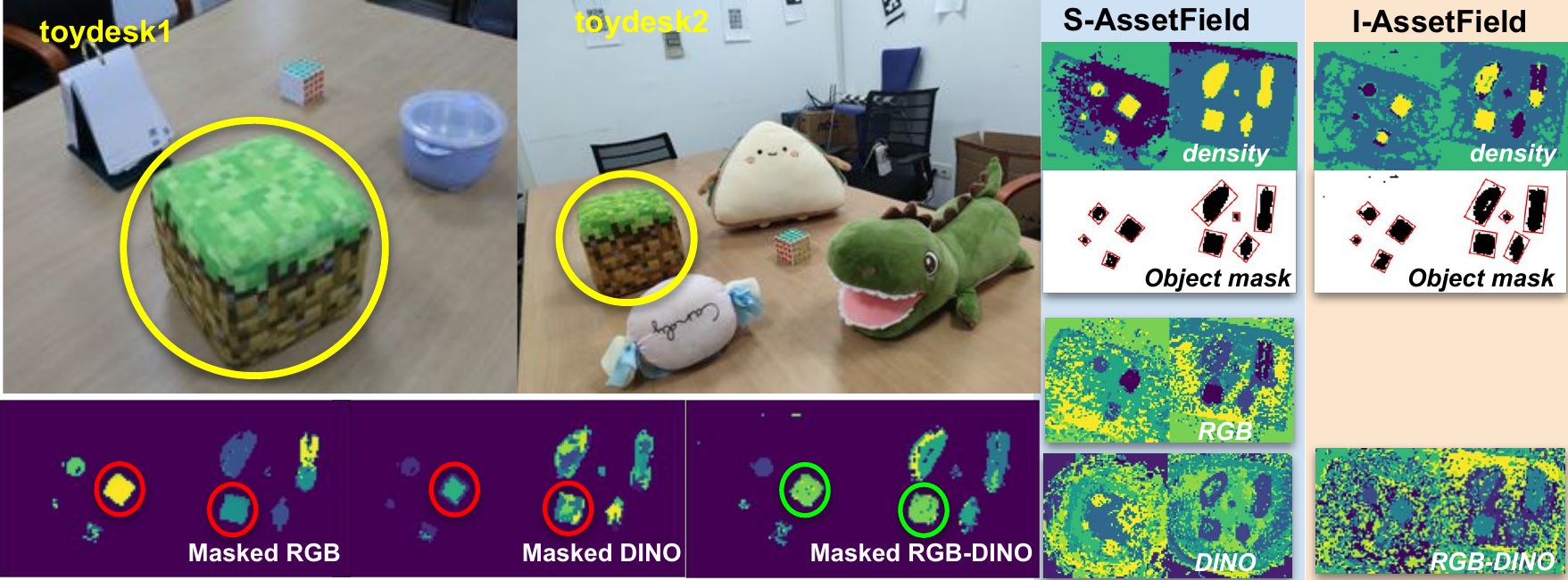}
	\vspace{-15pt}
	\caption{\small Multi-scene learning on the Toydesk dataset~\cite{yang2021objectnerf}. As real-world scenes usually exhibit noisier color and density features, we apply the object mask obtained from the density plane before categorization.
		The common object between scenes ({\color{yellow}yellow}) can be correctly clustered with I-AssetField's occupancy-guided RGB-DINO plane features ({\color{green}green}) whilst the independently modeled neural planes by S-AssetField fails ({\color{red}red}).}
	\label{fig:toydesk}
	\vspace{-15pt}
\end{figure}

\begin{figure*}[t!]
	\centering
	\includegraphics[width=\linewidth]{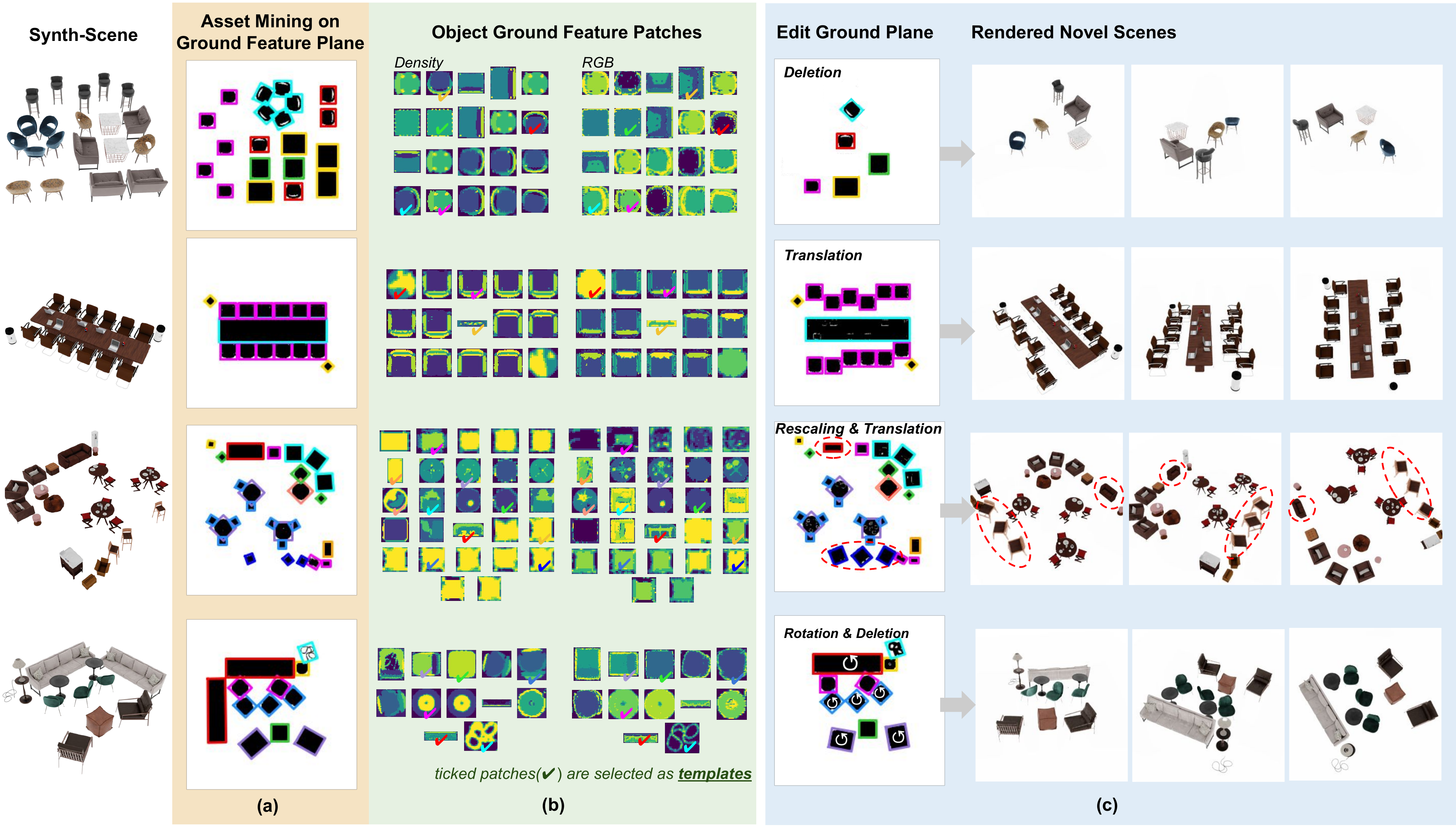}
	\vspace{-20pt}
	\caption{\small Results of assets mining and scene editing with \emph{I-AssetField} on synthetic scenes. (a) Our approach learns informative density and RGB-DINO ground feature planes that support object detection and categorization. (b) With joint training, an asset library can be constructed by storing ground \emph{feature plane patches} of the radiance field (we show label patches here for easy visualization). (c) The proposed ground plane representation provides an explicit visualization of the scene configuration, which can be directly manipulated by users. The altered ground feature planes are then fed to the global MLP renderer along with the shared vertical feature axes to render the novel scenes. Operations such as object removal, translation, rotation and rescaling are demonstrated on the right. }
	\vspace{-10pt}
	\label{fig:editing_results}
\end{figure*}

\noindent \textbf{Object Detection and Categorization.}
In Fig.~\ref{fig:tensorf_vs_groundplan} we already showed an example of the ground feature planes learned by AssetField compared to the $xy$-planes learned by TensoRF. 
While TensoRF produces noisy and less informative feature planes that is unfriendly for object discovery in the first place, AssetField is able to identify and categorize most of the scene contents, as shown in Fig.~\ref{fig:editing_results}~(b).
Furthermore, I-AssetField is more robust to vertical displacement, as shown in Fig.~\ref{fig:occupancy-guided}. On this scene variation, TensoRF/S-AssetField/I-AssetField achieves 35.873/36.358/36.452 in PSNR metric respectively on the test set.

Recall that I-AssetField is able to identify object analogues \emph{across different scenes}, to demonstrate such ability, we jointly model the two toy desk scenes from~\cite{yang2021objectnerf} by letting them share the same vertical feature axes and MLPs as described in Sec.~\ref{subsec:asset_library}. The inferred feature planes are showed in Fig.~\ref{fig:toydesk}.
Since the coordinate systems of these two scenes are not aligned with the physical world, we perform PCA~\cite{pca} on camera poses such that the $xy$-plane is expanded along the ground/table-top. However, we cannot guarantee their table surfaces are at the same height, meaning that vertical displacement among objects is inevitable. 
I-AssetField is able to infer similar RGB-DINO feature values for the common cube plush (yellow circle), whilst the independently learned RGB/DINO planes in S-AssetField are affected by the height difference.

\begin{figure}
	\centering
	\includegraphics[width=0.95\linewidth]{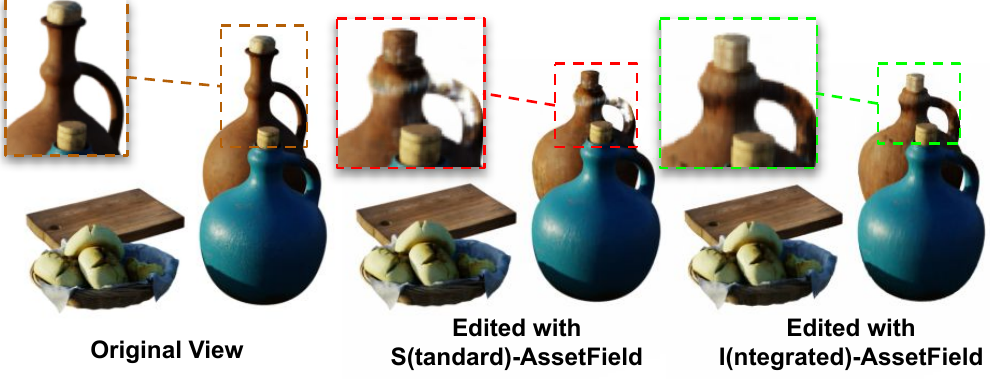}
	\vspace{-8pt}
	\caption{\small Density warping from the blue bottle to the region of the brown one. S-AssetField loses the structure of the brown bottle in terms of part semantics, while I-AssetField gives plausible editing result with appropriate structure transfer.}
	\vspace{-10pt}
	\label{fig:density-warping}
\end{figure}

\begin{figure}
	\centering
	\includegraphics[width=0.9\linewidth]{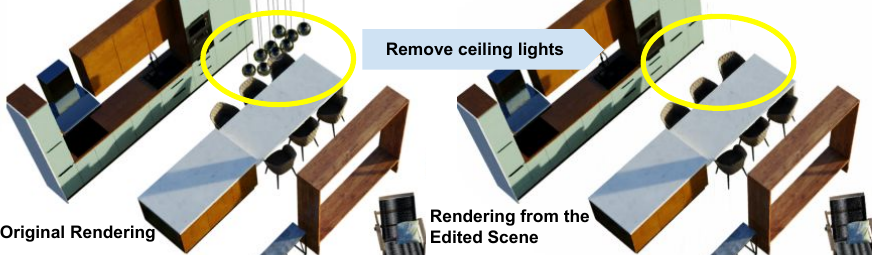}
	\vspace{-5pt}
	\caption{\small Expanding the 2D ground plane back to 3D feature grids, explicit control on full 3D space is allowed. We remove the ceiling light by setting the density grids as zero at the target region.}
	\vspace{-15pt}
	\label{fig:remove-ceiling}
\end{figure}

\begin{figure}
	\centering
	\includegraphics[width=\linewidth]{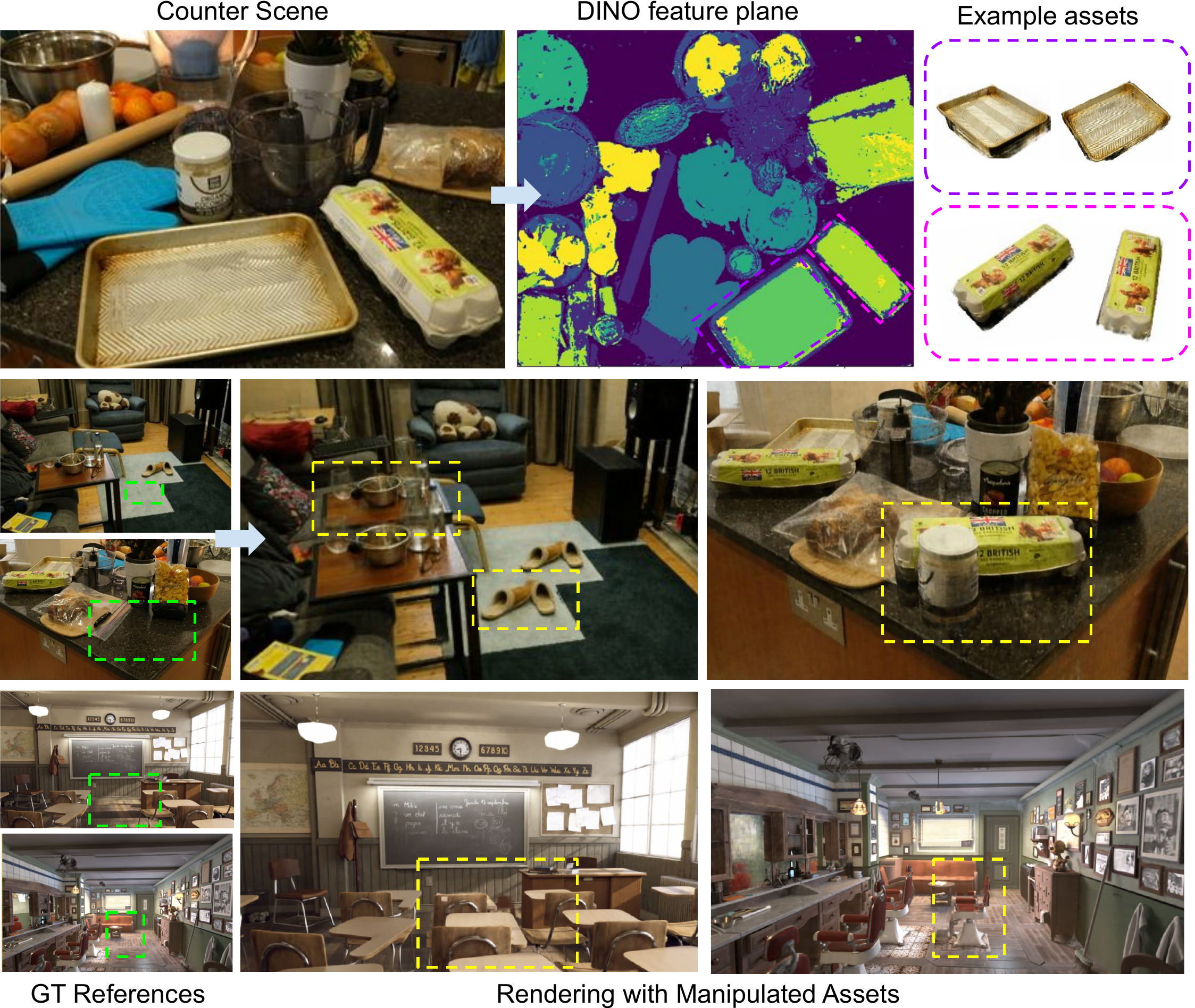}
	\vspace{-18pt}
	\caption{\small Example editings on real-world scenes~\cite{barron2022mip} and indoor scenarios~\cite{neff2021donerf}. We use RGB-DINO plane for assets discovery.
	}
	\vspace{-20pt}
	\label{fig:room-basic}
\end{figure}

\begin{figure}
	\centering
	\includegraphics[width=0.98\linewidth]{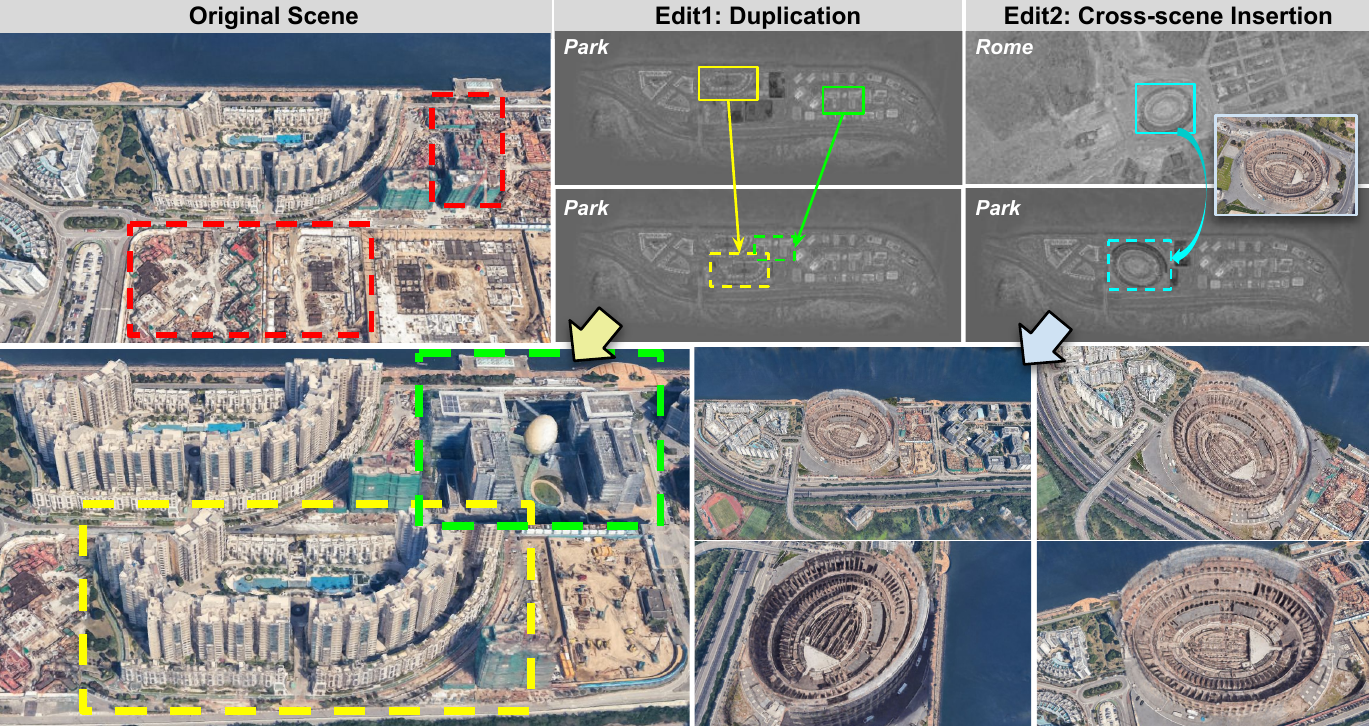}
	\vspace{-5pt}
	\caption{\small Editing two city scenes collected from Google Earth~\copyrightgoogle. AssetField is versatile where users can directly operate on the ground feature plane, supporting both within-scene and cross-scene editing with realistic rendering results.}
\vspace{-5pt}
\label{fig:cityscene}
\end{figure}

\noindent\textbf{Scene Editing.}
Techniques on 2D image manipulation can be directly applied to ground feature planes. Fig.~\ref{fig:editing_results} shows that AssetField supports a variety of operations, such as object removal, insertion, translation and rescaling. Scene-level reconfiguration is also intuitive by composing objects' density and color ground feature patches. 
In particular, I-AssetField associates the RGB-DINO field with space occupancy, producing more plausible warping results. Fig.~\ref{fig:density-warping} demonstrates a case of topology deformation, where the blue bottle's density field is warped to the region of the brown bottle, while keeping their RGB(-DINO) feature unchanged. Results show that I-AssetField successfully preserves object structure and part semantics, whereas S-AssetField fails to render the cork correctly.

Despite the convenience of ground feature plane representation, it does not directly support manipulating overlapping/stacked objects. However, one can expand the ground feature plane back to 3D feature grids with its pairing vertical feature axis, and control the scene in the conventional way as described in~\cite{liu2020neural}. An example is given in Fig.~\ref{fig:remove-ceiling}.

Fig.~\ref{fig:room-basic} shows AssetField's editing capability on real-world datasets~\cite{barron2022mip,yang2021objectnerf,neff2021donerf}. Additionally, on a self-collected city scene from Google Earth, we find a construction site and complete it with different nearby buildings (within-scene editing), even borrow Colosseum from Rome (cross-scene editing). Results are shown in Fig~\ref{fig:cityscene}. The test view PSNR for the original scene is NeRF/TensoRF/S-AssetField/I-AssetField: 24.55/27.61/ 27.54/ 27.95.

\begin{figure}
	\centering
	\includegraphics[width=\linewidth]{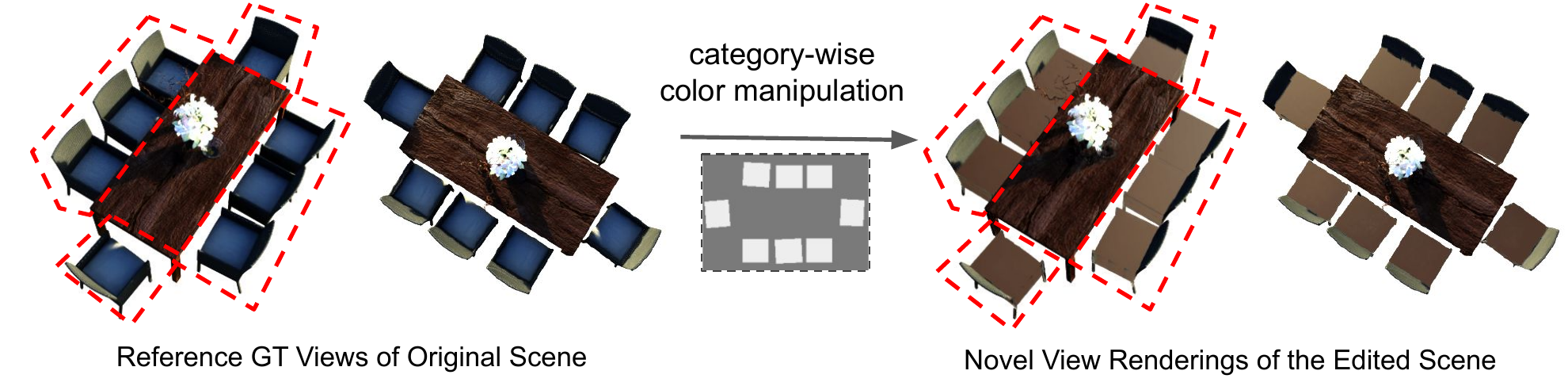}
	\vspace{-20pt}
	\caption{\small We apply batch-wise color changing for all instances of the chair, by replacing the template RGB feature map solely.}
	\label{fig:category}
	\vspace{-15pt}
\end{figure}

\begin{figure}
	\centering
	\includegraphics[width=0.9\linewidth]{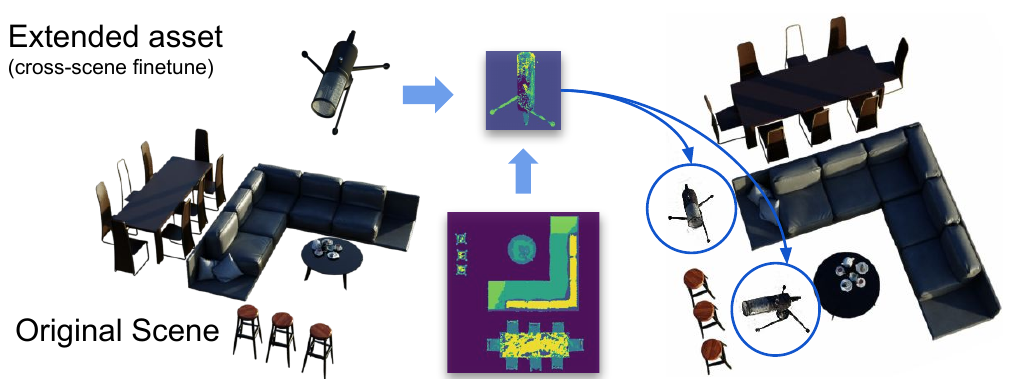}
	\vspace{-10pt}
	\caption{\small We expand the asset library from the living room with the newly included assets \emph{mics} from~\cite{mildenhall2020nerf}. The template of mics is in the shared latent space with the living room and can thus naturally composed together for rendering.}
	\label{fig:cross-scene}
	\vspace{-10pt}
\end{figure}

\begin{figure}
	\centering
	\includegraphics[width=0.9\linewidth]{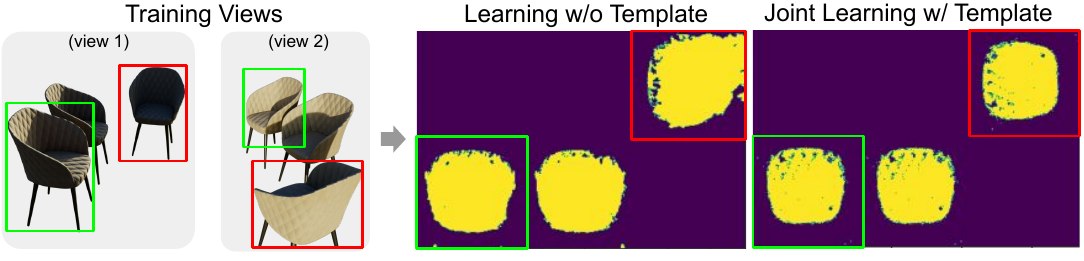}
	\vspace{-10pt}
	\caption{\small Feature plane refinement. The object template, when trained among all instances within the scene, produces more accurate feature map compared to the isolated ones.}
	\label{fig:template}
	\vspace{-20pt}
\end{figure}

\noindent \textbf{Group Editing and Scene Reconfiguration.}
Recall that a template object can be selected for each asset category to substitute all its instances in the scene (on the ground feature planes). Consequently, we are allowed to perform group editing like changing the color of a specific category as depicted in Fig.~\ref{fig:category}. 
Scene-level reconfiguration is also intuitive, where users can freely compose objects from the \emph{asset library} on a neural `canvas' to obtain a set of new ground feature planes, as demonstrated in Fig.~\ref{fig:cross-scene}.
The environments or containers~(\eg the floor or an empty house) can also be considered as a special asset category, where small objects~(\eg furniture) can be placed into the container to deliver immersive experience. The final scene can be composited with summed density value and weighted color, as has been discussed in~\cite{Tang2022CompressiblecomposableNV}.

\smallskip
\noindent \textbf{Template Refinement.} 
Grid-based neural fields are sensitive to training views with insufficient point supervision, leading to noisy and inaccurate feature values.
Appearance differences caused by lighting variation, occlusion, etc., interferes the obtaining of a clean template feature patch. 
An example can be found in Fig.~\ref{fig:template}.
Due to imbalanced training view distribution, the chair in the corner receives less supervision, resulting in inconsistent object feature patch within a category. 
Such issue can be alleviated with a following-up \emph{template refinement} step.
With the inferred scene layout and the selected object templates (Sec.~\ref{subsec:asset_grouping}).
We propose to replace all instances $p\in\mathcal{P}$ with their representative category template $\hat{p}$ and optimize this set of feature patches to reconstruct the scene instead of the full ground planes. 
Consequently, the template feature patch integrates supervisions from all instances in the scene to overcome appearance variations and sparse views. 

%% file: sections/5_conclusion.tex
\section{Discussion and Conclusion}
\label{sec:conclusion}
We present AssetField, a novel framework that mines assets from neural fields. 
We adopt a ground feature plane representation to model scene density, color and semantic fields, on which
assets mining and grouping can be directly conducted. 
The novel occupancy-guided RGB-DINO feature plane enables cross-scene asset grouping and the construction of an expandable neural  asset library, enabling
a variety of intuitive scene editing at object-, category- and scene-level.
Extensive experiments are conducted to show the easy control over multiple scenes and the realistic rendering results given novel scene configurations.
However, AssetField still suffer from limitations like:
separating connected objects in the scene;
handling stacked/overlapped objects; and performing vertical translations.
Rendering quality might also be compromised due to complex scene background in real-world. 
More limitations are discussed in the supplementary.
We believe the proposed representation can be further explored for the manipulation and construction of large-scale scenes,~\eg, by following floorplans or via a programmable scheme like \emph{procedural modeling}.